\documentclass{article}

\usepackage{arxiv}
\usepackage{amsfonts}

\usepackage[utf8]{inputenc} 
\usepackage[T1]{fontenc}    
\usepackage{hyperref}       
\hypersetup{colorlinks=true, linkcolor=blue, filecolor=blue,  urlcolor=blue, citecolor=blue }
\usepackage{url}            
\usepackage{booktabs}       
\usepackage{amsfonts}       
\usepackage{nicefrac}       
\usepackage{microtype}      
\usepackage{lipsum}		
\usepackage{graphicx}
\usepackage{natbib}
\usepackage{doi}
\usepackage{float}
\usepackage{longtable}
\usepackage{amsmath}

\title{A Comparative Evaluation of Transformer Models for De-identification of Clinical Text Data}

\author{ 
    \href{https://orcid.org/0000-0002-5429-5233}{\includegraphics[scale=0.06]{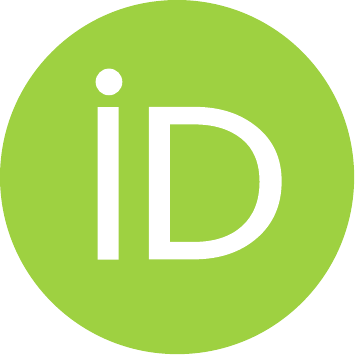}\hspace{1mm}Christopher Meaney}\\
	Department of Family and Community Medicine\\
	University of Toronto\\
	Toronto, Ontario, Canada \\
	\texttt{christopher.meaney@utoronto.ca}
	\And
	\href{https://orcid.org/0000-0001-7410-5380}{\includegraphics[scale=0.06]{orcid.pdf}\hspace{1mm}Wali Hakimpour}\\
	Department of Family and Community Medicine\\
	University of Toronto\\
	Toronto, Ontario, Canada \\
	\texttt{ahmad.hakimpour@mail.utoronto.ca}
	\And
	\href{https://orcid.org/0000-0002-8165-4959}{\includegraphics[scale=0.06]{orcid.pdf}\hspace{1mm}Sumeet Kalia}\\
	Department of Family and Community Medicine\\
	University of Toronto\\
	Toronto, Ontario, Canada \\
	\texttt{sumeet.kalia@utoronto.ca}
	\And
	\href{https://orcid.org/0000-0002-5506-084X}{\includegraphics[scale=0.06]{orcid.pdf}\hspace{1mm}Rahim Moineddin}\\
	Department of Family and Community Medicine\\
	University of Toronto\\
	Toronto, Ontario, Canada \\
	\texttt{rahim.moineddin@utoronto.ca}
}

\renewcommand{\headeright}{}
\renewcommand{\undertitle}{Technical Report}

\hypersetup{
pdftitle={Tranformer Models for Clinical Text De-identification},
pdfsubject={stat.AP},
pdfauthor={Christopher Meaney, Wali Hakimpour, Sumeet Kalia, Rahim Moineddin},
pdfkeywords={Clinical Text Data, De-identification, Transformer, Deep Learning, Natural Language Processing},
}

\begin{document}
\maketitle

\begin{abstract}

\textbf{Objective:} To comparatively evaluate several transformer model architectures at identifying protected health information (PHI) in the i2b2/UTHealth 2014 clinical text de-identification challenge corpus.\\

\textbf{Methods:} The i2b2/UTHealth 2014 corpus contains N=1304 clinical notes obtained from N=296 patients. Using a transfer learning framework, we fine-tune several transformer model architectures on the corpus, including: BERT-base, BERT-large, ROBERTA-base, ROBERTA-large, ALBERT-base and ALBERT-xxlarge. During fine-tuning we vary the following model hyper-parameters: batch size, number training epochs, learning rate and weight decay. We fine tune models on a training data set, we evaluate and select optimally performing models on an independent validation dataset, and lastly assess generalization performance on a held-out test dataset. We assess model performance in terms of accuracy, precision (positive predictive value), recall (sensitivity) and F1 score (harmonic mean of precision and recall). We are interested in overall model performance (PHI identified vs. PHI not identified), as well as PHI-specific model performance. \\

\textbf{Results:} We observe that the ROBERTA-large models perform best at identifying PHI in the i2b2/UTHealth 2014 corpus, achieving >99\% overall accuracy and 96.7\% recall/precision on the held-out test corpus. Performance was good across many PHI classes; however, accuracy/precision/recall decreased for identification of the following entity classes: professions, organizations, ages, and certain locations.\\

\textbf{Conclusions:} Transformers are a promising model class/architecture for clinical text de-identification. With minimal hyper-parameter tuning transformers afford researchers/clinicians the opportunity to obtain (near) state-of-the-art performance. 

\end{abstract}


\section{Introduction}

There exist several international standards which govern the use of patient health data in particular jurisdictions, for example: PHIPA in Canada (\href{https://www.ontario.ca/laws/statute/04p03}{Personal Health Information Protection Act Health}{}), GDPR in the European Union (\href{https://eur-lex.europa.eu/legal-content/EN/TXT/?uri=CELEX:32016L0680}{General Data Protection Regulation} ) and HIPAA in the United States (\href{https://www.hhs.gov/sites/default/files/ocr/privacy/hipaa/understanding/coveredentities/De-identification/hhs_deid_guidance.pdf}{Health Insurance Portability and Accountability Act}). From the perspective of clinical text data for secondary research purposes, the HIPAA safe harbor standard is most widely adopted. The safe harbor definition specifies which entities must be identified and redacted/substituted/augmented from the original clinical text documents such that it is suitable for use in research. These entities include: names, locations, dates, phone/fax numbers, emails, URLs, facial imagery, and various unique identifiers (medical record numbers, IP addresses, social insurance numbers, etc.). De-identification of record-level PHI in clinical text data pertains to both patient identifiers, as well as identifiers related to a patients’ care providers, household members, relatives and/or employers.\\

Suitable de-identification of clinical text data involves two modular steps: 1) identification of PHI entities in the clinical text document, and 2) redaction, tag insertion or substitution of identified PHI entities. Under the HIPAA safe harbor definition, step one involves the development of algorithms capable of processing large amounts of clinical text data, to identify digital character sequences corresponding HIPAA defined PHI entities. It is imperative that any proposed algorithm identify named entities (PHI) with both high sensitivity (ensuring few PHI elements are missed in the data transformation process) and high positive predictive value (ensuring the preservation of medically relevant information, semantic readability of the text document, and avoiding over-scrubbing). Given an accurate algorithm for identification of PHI in clinical text, the next step in development of a successful de-identification system involves suitable redaction, tag insertion, or substitution of PHI entities from the text passages. Redaction and tag insertion are commonly employed in practice, but viewed as inferior solutions, as missed PHI entities (which occur when an identification algorithm does not have 100\% sensitivity) remain in the clinical text. A preferable solution involves the replacement of identified PHI entities using “realistic surrogates” – which are often (randomly) drawn/sampled from external data sources (e.g. gender specific name-banks, location banks, date-shifting, etc.). In this study we focus on the evaluation of machine learning algorithms for the identification of PHI elements, and do not further explore strategies for masking identified PHI elements in clinical text data \citep{yogarajan2020review}.\\  

The objective of this study is to fine-tune and comparatively evaluate novel transformer architectures, which have exceled at various natural language processing tasks, for use in clinical text de-identification. We fine-tune several transformer architectures (i.e. BERT, ROBERTA, and ALBERT) on the i2b2/UTHealth 2014 clinical text de-identification dataset; and subsequently evaluate model performance on a held-out test set. We assess model performance on the overall PHI de-identification task (i.e. can the algorithm identify PHI vs non-PHI tokens), as well as the algorithms ability to identify particular PHI sub-classes (e.g. names, dates, locations, etc.). We discuss differences in model performance as a function of hyper-parameter configurations (batch size, number training epochs, learning rate, weight decay). Further, we discuss differences in training/data-processing time and computational requirements for each of the models.\\ 

\section{Methods}

\subsection{Data Source}

In this study we use the i2b2/UTHealth 2014 longitudinal clinical narrative corpus for comparatively evaluating various transformer architectures for de-identification of clinical PHI. The corpus is commonly utilized in evaluating clinical de-identification systems and its annotation scheme is described in \citet{stubbs2015annotating}. Top performing models from the i2b2/UTHealth 2014 PHI de-identification data challenge are described in \citet{stubbs2015automated}. The i2b2/UTHealth 2014 dataset is publicly available for research purposes and can be obtained from the N2C2 research consortia: \url{https://portal.dbmi.hms.harvard.edu/projects/n2c2-nlp/}. \\

\subsection{Data Description, Annotation Guidelines and Tag-Set}

The i2b2/UTHealth 2014 corpus contains longitudinal clinical records from N=296 patients; the corpus consists of N=1,304 clinical notes. Clinically, all N=296 patients are diagnosed with diabetes. Further, the notes can be divided into three substrata: 1) patients with comorbid coronary artery disease (first and subsequent visits/notes), 2) patients without coronary artery disease at baseline, who go on to develop the condition on subsequent visits, and 3) patients with no mention of coronary artery disease in any of their notes/visits. \\

The corpus consists of a total of N=805,118 tokens, with N=28,872 (3.58\%) of these tokens representing PHI. On average 22.14 PHI tokens exist in each clinical note (which itself is 617.4 tokens in length). In terms of particular PHI classes: dates (N=12,847), names (N=7,348), locations (N=4,580), ages (N=1,997) and medical record numbers (N=1,033) are some of the most frequently occurring entity types. The i2b2/UTHealth 2014 researchers chose to annotate their clinical records against a “risk-averse” HIPAA interpretation (i.e. they included all HIPAA PHI categories, plus additional sensitive linguistic entity types, such as: professions and organizations). PHI entities in the annotated dataset included: \\

\begin{itemize}
	\item Name
	\item Patients
	\item Doctors
	\item Usernames
	\item Profession
	\item Location
	\item Hospital
	\item Organization
	\item Street
	\item City
	\item State
	\item Country
	\item Zip code
	\item Other
	\item Age
	\item Date
	\item Contact Information
	\item Phone
	\item Fax
	\item Email
	\item URL
	\item IP address
	\item Identification Numbers
	\item Social security numbers
	\item Medical record numbers
	\item Health plan
	\item Accounts
	\item License
	\item Vehicle identification number
	\item Device
	\item Biological identifiers
	\item Other ID numbers
\end{itemize} 

A complete description of the i2b2/UTHealth annotation guidelines is given in \citet{stubbs2015annotating}. A description of corpus statistics, including a detailed breakdown of the frequency of particular PHI tokens is provided in Table 2 of \citet{stubbs2017identification}.\\

\subsection{Text Pre-Processing}

The i2b2/UTHealth 2014 dataset is provided to researchers as a collection of XML files. Each XML file represents a single clinical encounter, and is hierarchically structured into two sub-components: 1) “CDATA” which contains the digital character sequences (UTF8 encoded) corresponding to each clinical text document, and 2) “TAGS” which describe the character spans (i.e. start and stop locations in the clinical text), where PHI entities have been identified as part of the annotation process. \\

We write an R program (using base R string processing functions) to tokenize the clinical text data, splitting the digital character sequences into linguistic units (i.e. tokens/words). Using the tag-set information (and corresponding PHI character spans) we identify which tokens are PHI and non-PHI. Further, PHI tokens are tagged using a common “BIO-tagging” methodology (often employed in named entity recognition problems in natural language processing). This BIO methodology allows for multi-token PHI entities (e.g. locations: “New York City”; organizations: “Partners Healthcare”; professions “Orthopedic Surgeon”, etc.) to be tagged as 1) (B)eginning, 2) (I)nside, or 3) (O)utside of the PHI tag-set. In total, we observe 41 BIO-tags in this study (see Appendix A). When fine-tuning models we use the BIO-tags; however, when evaluating models we collapse the B/I-tags and focus on 21 tags (20 PHI types, and 1 non-PHI type label). \\

Token-to-tag alignment is a challenging task. Starting from N=1304 clinical notes we were not able to align all tokens/tags for every clinical document in the corpus. In total, we were able to align 1223 of the documents in the i2b2/UTHealth 2014 corpus - and these aligned documents will be used in training (fine-tuning), validation and testing of our transformer models. \\

\subsection{Transformer Models}

The seminal paper on transformers is \citet{vaswani2017attention}. The original transformer model architecture from \citet{vaswani2017attention} is illustrated in Figure 1 below. The transformer model consists of an encoder/decoder architecture (frequently observed when processing sequential data in natural language processing settings). The encoder module processes word/token embeddings and (relative/absolute) positional embeddings arising from a source/input sequence. The encoder consists of four sub-modules: 1) a multi-head self-attention module, 2) an initial residual connection and layer normalization module, 3) a point-wise feed-forward neural network, and 4) a final residual connection and layer normalization module. The decoder module processes information from two sources: word/token embeddings and positional embeddings related to the output/shifted sequence, as well as intermediate hidden representations generated by the encoder. The decoder sub-module is slightly more complex than the encoder, consisting of six sub-modules: 1) (masked) multi-head attention over the output embeddings, 2) an initial set of residual connections and layer normalization module over the output embeddings, 3) cross-attention over the encoded input sequence, 4) an intermediate residual connection and layer normalization module, 5) a point-wise feed-forward neural network, and 6) a final residual connection and layer normalization module. The outputs of the decoder module are finally projected onto the vocabulary/targets using a linear transformation, and a final softmax transformation is used to map scores/predictions over the vocabulary/targets onto a probabilistic scale. The transformer modules are hierarchically stacked (Nx times), yielding a deep transformer architecture. \\

\begin{figure}[H]
	\centering
	\includegraphics[width=0.5\textwidth]{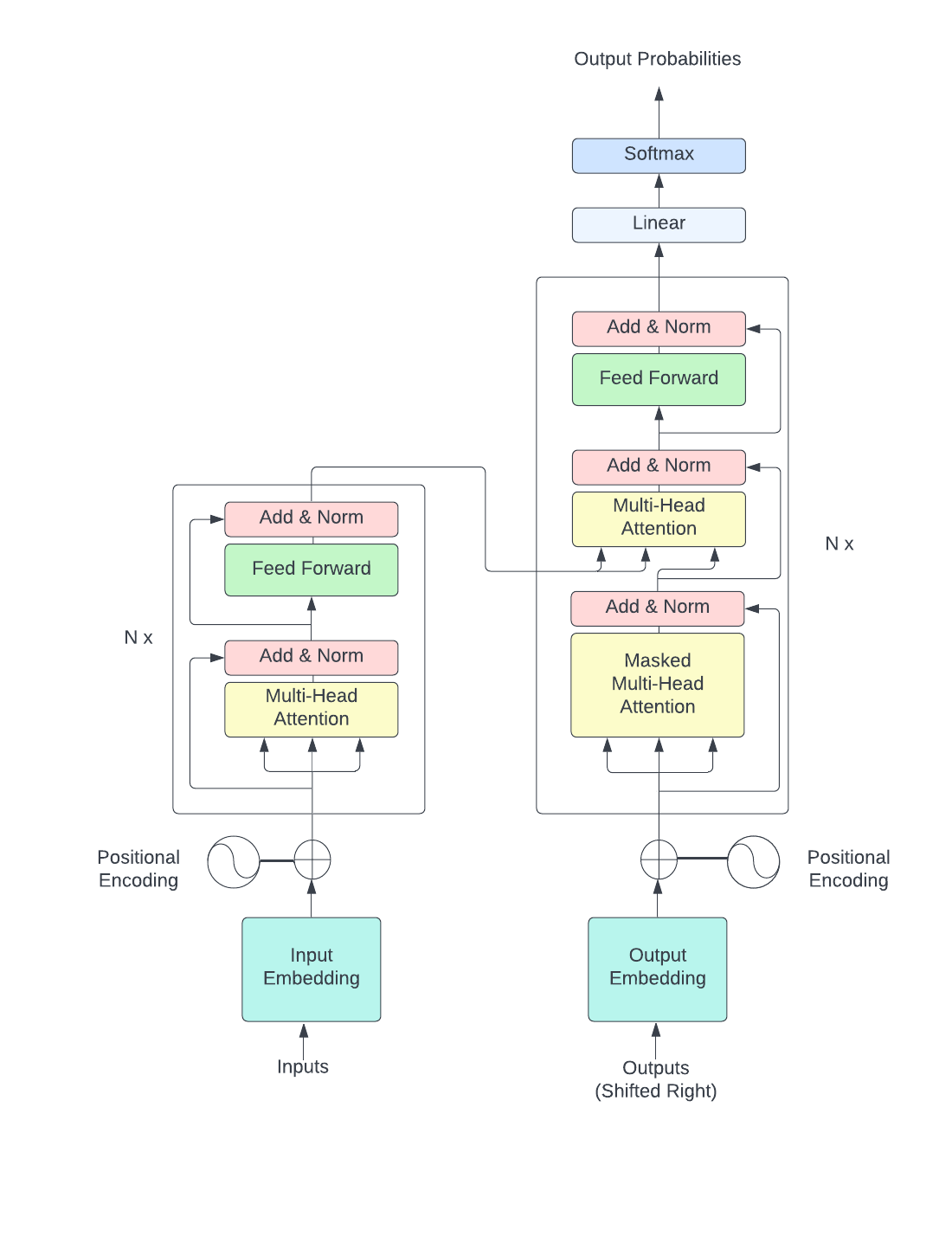}
	\caption{Transformer model from \citet{vaswani2017attention}.}
	\label{fig:fig1}
\end{figure}

The core components of the transformer architecture are: 1) word/token embeddings along with positional embeddings, 2) encoder self-attention, encoder/decoder cross attention, and decoder masked self-attention, 3) residual connections and layer normalization, and 4) point-wise feed-forward neural network layers. Below we briefly discuss each component. \\

Words/language-elements are inherently discrete. Rather than work with digital character sequences, or a sequence of discrete word/token symbols, the transformer model ingests words in the form of embeddings (i.e. continuous/numeric vector representations used to capture multiple degrees of semantic and syntactic similarity between the input words/tokens). Depending on the downstream task under consideration, there may not exist a particular embedding/representation for all words in a corpus (i.e. certain words/tokens may be “out of vocabulary” relative to the initially chosen training corpus). For example, in the i2b2/UTHealth 2014 corpus there exist highly specific medical terms/nomenclature, rare medical abbreviations, and frequent mis-spellings – hence, transformer models use a combination of word/token level embeddings (to represent commonly occurring words/tokens) which are expected to exist across corpora, as well as character-level/byte-level encodings which allow numeric vector representations of rare/unseen words to be composed from their constituent sub-components (e.g. morphemes or even individual characters).\\

Transformer models (unlike recurrent networks for sequential data) process an entire sequence (or its embedded vector representation) as a single unit, and as such, do not naturally encode word/token ordering in their representation. To circumvent this issue, transformers inject relative/absolute ordering information into word/token embeddings using an additional positional encoding. The positional embeddings defined in the original transformer paper are sine/cosine functions \citep{vaswani2017attention}. \\

Stacked (multi-headed) attention layers are a defining feature of transformer architectures. An attention module creates a new attention weighted (context specific) vector representation, of a given word/token of the input sequence. For a given token ($x_i$) represented as a d-dimensional embedding vector we construct a query vector ($q_i = x_i\times Q$), key vector ($k_i = x_i\times  K$) and value vector ($v_i = x_i \times V$). The non-symmetric attention that a given token ($x_i$) shows another token ($x_j$) can be expressed by the inner product of the query vector and the key vector: $a_{ij} = q_i \times k_j^T$. The $\sqrt{d_k}$ term divides attention weights by the dimension of the embeddings (empirically, stabilizing gradient calculations); and the softmax() function converts attention weights to a probabilistic scale. A single attention layer/module can be represented using the matrix calculations given below. In standard transformer architectures multi-headed attention modules are employed (i.e. independently layering multiple self-attention modules), allowing different attention heads to provide different contextual representations for a given word/token. In transformers we also employ cross-attention between encoder/decoder stacks of the model. And the decoder also uses masked attention when ingesting outputs (as to not attend to future contexts). Below, \{Q,K,V\} represent learnable weight matrices from the attention modules. \\

\begin{center}
\begin{equation*}
\text{Attention}(Q,K,V) = \text{softmax}\left(\frac{Q\times K^T}{\sqrt{d_k}}\right)\times V
\end{equation*}
\end{center}

\begin{figure}[H]
	\centering
	\includegraphics[width=0.85\textwidth]{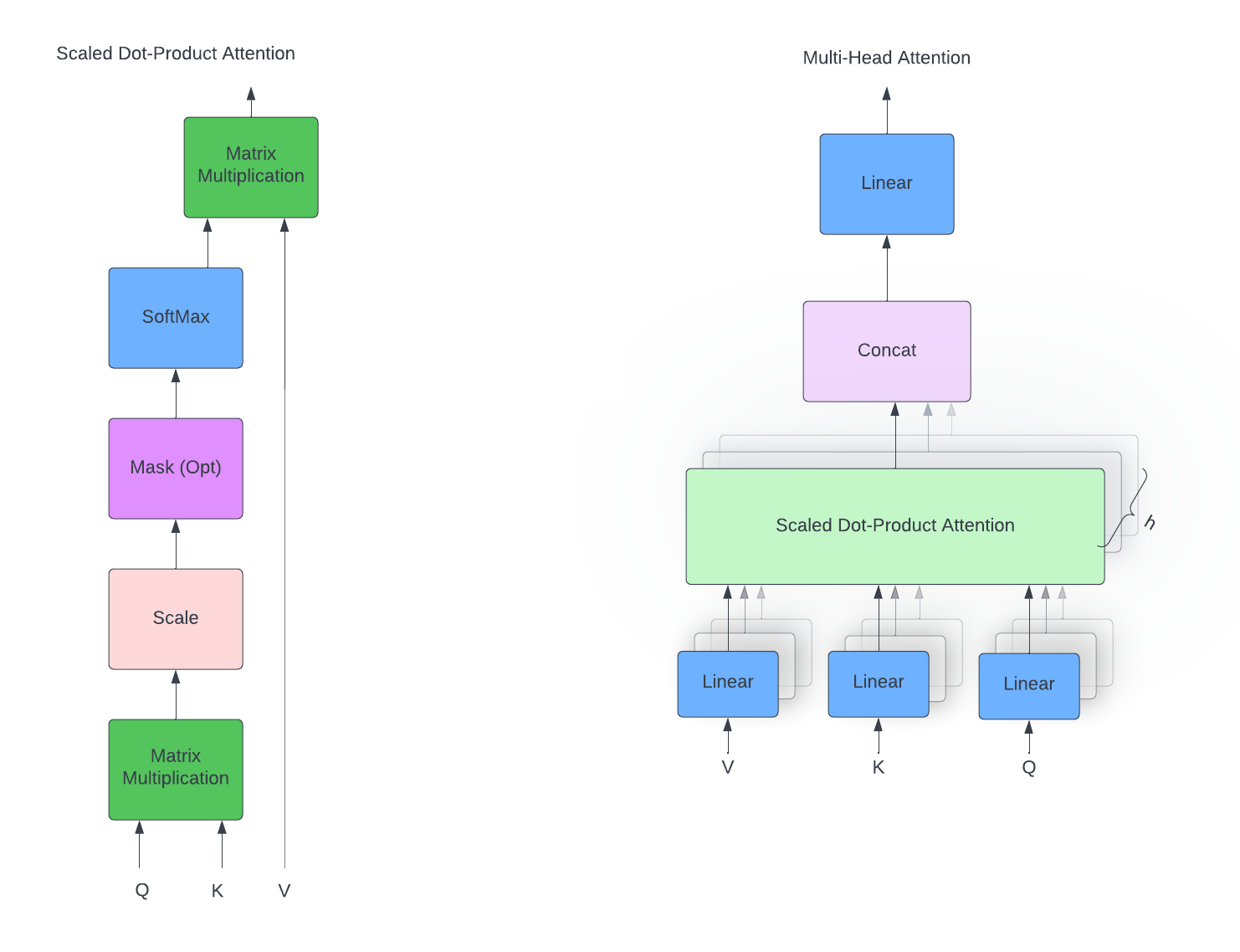}
	\caption{A visual representation of scaled dot-product attention, and multi-head attention from \citet{vaswani2017attention}.}
	\label{fig:fig2}
\end{figure}

Point wise feed-forward networks (with Gaussian error linear unit – GELU - activation functions) are used to generate non-linear representations of the contextualized embeddings. Residual connections and layer normalization are used to stabilize training the deep transformer architecture \citep{ba2016layer}. \\

\subsection{BERT, ROBERTA and ALBERT Pre-training Strategy and Model Architectures}

In this study we comparatively evaluate several BERT-variant models (i.e. bidirectional encoder representations from transformer) for identifying PHI in clinical text data from the i2b2/UTHealth 2014 corpus. The BERT-variant models differ from the original transformer model architecture described in \citet{vaswani2017attention} in that they use only the encoder stack to generate a deep representation of the input sequence (i.e. no decoder is used in BERT-variant models). The core components of the model include: word/token embeddings, positional embeddings, and stacked transformer modules (with self-attention, residual connections, layer normalization, and a point-wise feed-forward network). The aforementioned modules are utilized in all models considered (i.e. BERT, ROBERTA, and ALBERT). \\

While sharing many similar attributes, the BERT, ROBERTA and ALBERT models differ in subtle/nuanced manners. For example, the pre-training loss/objective varies across each of the models; hence the output layer of each model (during pre-training) will differ to reflect the different objective. Further, the particular pre-training corpus used may also differ across models. Finally, there exist some architectural differences that distinguish the models. Below we discuss some of the nuanced differences associated with each of the BERT, ROBERTA, and ALBERT models used in this study. In Appendices B-G  we provide a precise model description for each of the BERT-base, BERT-large, ROBERTA-base, ROBERTA-large, ALBERT-base and ALBERT-xxlarge models used in this study. \\

\subsubsection{Bidirectional Encoder Representations from Transformers (BERT)}

The seminal BERT (bi-directional encoder representations form transformers) model is described in \citet{devlin2018bert}. The BERT model is pre-trained on the Toronto/Google books corpus (16Gb), plus a 3.3-billion-word corpus extracted from scraping Wikipedia pages. The model is pre-trained using 1) a masked language modelling loss, and 2) a next sentence prediction loss. Aspects of both the BERT-base and BERT-large model are described in Table 1, below. A complete description of the BERT-base and BERT-large model architecture, along with associated parameter counts are given in Appendix B and Appendix C, respectively. The Huggingface implementation of BERT-base/BERT-large are used for fine-tuning on the i2b2/UTHealth de-identification corpus: \url{https://huggingface.co/transformers/model_doc/bert.html} \\

\subsubsection{Robustly Optimized BERT Pre-Training Approach (ROBERTA)}

The seminal ROBERTA model and pre-training strategy is described in \citet{liu2019roberta}. The ROBERTA model architecture is the same as the BERT model architecture. The model pre-training corpus for ROBERTA is much larger than the one used in the original BERT paper. ROBERTA is pre-trained with respect to a 160Gb text corpus (i.e. 16Gb corpus from BERT, plus an additional 144Gb of text data scraped from the Internet). The ROBERTA model is pre-trained using a different loss function than was used in the original BERT paper. ROBERTA continues to use the masked language model loss function; however, removes the next sentence prediction loss. Further, ROBERTA employs a dynamic masking approach such that different words/tokens are masked over different training epochs (whereas, BERT employs a simpler static masking approach). Finally, the ROBERTA model is trained for many more epochs on a much larger hardware stack than was used in BERT training. Details regarding ROBERTA-base and ROBERTA-large architectural decisions are given in Table 1 below. A complete description of the ROBERTA-base and ROBERTA-large model architecture, along with associated parameter counts are given in Appendix D and Appendix E, respectively. We use the Huggingface implementation of ROBERTA-base and ROBERTA-large for fine-tuning on the i2b2/UTHealth de-identification corpus: \url{https://huggingface.co/transformers/model_doc/roberta.html} \\

\subsubsection{A Lite BERT for Self-Supervised Learning of Language Representations (ALBERT)}

The seminal ALBERT model was introduced in \citet{lan2019albert}. The authors attempted to address the issue that enhanced transformer model performance was observed to be associated with larger/deeper models; however, side-effects associated with utilizing these larger models included increased hardware requirements, memory utilization, and training times. The paper proposed two architectural changes relative to the initial BERT paper, namely: 1) the use of factorized embedding layers for representing input word/tokens, and 2) cross-layer parameter sharing/tying. Further, the ALBERT pre-training objective included a masked language modelling loss; however swapped out the next sentence prediction task in favour of a novel sentence order prediction loss. Details regarding ALBERT-base and ALBERT-xxlarge architectural decisions are given in Table 1 below. A complete description of the ALBERT-base and ALBERT-xxlarge model architecture, along with associated parameter counts are given in Appendix F and Appendix G, respectively. We use the Huggingface implementation of ALBERT-base and ALBERT-xxlarge for fine-tuning on the i2b2/UTHealth de-identification corpus: \url{https://huggingface.co/transformers/model_doc/albert.html} \\

\begin{table}[H]
	\caption{Model size and architectural characteristics of BERT, ROBERTA and ALBERT models.}
	\centering
    \scalebox{0.75}{
	\begin{tabular}{lccccccccccc}
		\toprule
		Model     & Size & Number & Number & Vocabulary & Embedding & Attention & Feed Forward & Dropout & Activation & Parameter & Objective \\
		          & $\text{(Gb)}^{ a}$ & Parameters & Layers & Size & Size & Size & Network Size & & $\text{Function}^{ b}$ & Sharing & $\text{Function}^{ c}$ \\
		\midrule
		Bert-Base       &  0.42 & 108,923,177 & 12 & 30,522 & 768	& 768 &	3072 &	0.1	& GELU & False & MLM/NSP \\
		Roberta-Base    &  0.47 & 124,086,569 & 12 & 50,265 & 768	& 768 &	3072 & 0.1 & GELU &	False & MLM \\
		Albert-Base     &  0.04 & 11,124,521 & 12 & 30,000 & 128	& 768 &	3072 &	0.0	& GELU & True &	MLM/SOP \\
		Bert-Large      &  1.24 & 334,134,313 & 24 & 30,522 & 1024 & 1024 & 4096 & 0.1 & GELU & False & MLM/NSP \\	Roberta-Large   &  1.32 & 354,352,169 & 24 & 50,265 & 1024 & 1024	& 4096 & 0.1 & GELU & False & MLM \\
		Albert-XXLarge  &  0.79 & 205,982,249 & 12 & 30,000 & 128	& 4096 & 16,384 & 0.0 & GELU & True & MLM/SOP \\
		\bottomrule
	\end{tabular}
	}
	\label{tab:table1}
\end{table}
\footnotesize{$^a$ Model size (Gb) denotes the size of the HuggingFace implementation of the particular transformer model binary file.}\\
\footnotesize{$^b$ GELU: Gaussian Error Linear Unit.}\\
\footnotesize{$^c$ MLM: Masked Language Modelling; NSP: Next Sentence Prediction; SOP: Sentence Order Prediction.}\\

\subsection{Design for Comparatively Evaluating Transformer Architectures for Clinical Text De-identification}

We comparatively evaluate six transformer models for identification of PHI entities in the i2b2/UTHealth 2014 corpus: BERT-base, BERT-large, ROBERTA-base, ROBERTA-large, ALBERT-base, and ALBERT-xxlarge. When fine-tuning the models, our goal is to choose the largest batch size possible such that the data/model fit into GPU memory. We specify a hyper-parameter grid for fine-tuning over: number of epochs, learning rate and weight decay. In table 2 below we expand on the hyper-parameter configuration over which we search for an optimal model. We randomly divide the i2b2/UTHealth 2014 corpus into independent training/validation/test datasets (using a 50:10:40 random split). We fine tune models on the training dataset, and use the F1-metric to identify the best performing model with each of the six model classes using an independent validation dataset (i.e. BERT-base, BERT-large, ROBERTA-base, ROBERTA-large, ALBERT-base, and ALBERT-xxlarge). We assess generalization performance using precision, recall, F1-score and accuracy; each assessed on a held-out test set. \\

\begin{table}[H]
	\caption{Hyper-parameter configurations for fine-tuning transformer models on the i2b2/UTHealth 2014 de-identification dataset.}
	\centering
    \scalebox{0.90}{
	\begin{tabular}{lcccc}
		\toprule
	    Model & Batch Size & Number Epochs & Learning Rate & Weight Decay \\
	    \midrule
        BERT-Base	& 35 &	(10, 20, 30, 40, 50) & (0.0001, 0.00001) & (0.0, 0.01, 0.025) \\
        ROBERTA-Base & 35 & (10, 20, 30, 40, 50) & (0.0001, 0.00001) & (0.0, 0.01, 0.025) \\
        ALBERT-Base & 35 & (10, 20, 30, 40, 50)	& (0.0001, 0.00001) & (0.0, 0.01, 0.025) \\
        BERT-Large & 12 & (10, 20, 30, 40, 50) & (0.0001, 0.00001) & (0.0, 0.01, 0.025) \\
        ROBERTA-Large & 12 & (10, 20, 30, 40, 50, 75, 100) &	(0.0001, 0.00001, 0.000001) & (0.0, 0.01, 0.025, 0.05, 0.10) \\
        ALBERT-XXLarge & 6 & (10, 20, 30, 40, 50) & (0.0001, 0.00001) & (0.0, 0.01, 0.025) \\
    \bottomrule
	\end{tabular}
	}
	\label{tab:table2}
\end{table}

\subsection{Computational Environment and Data Scientific Software Stack}

Computations were performed on the MIST supercomputer at the SciNet HPC Consortium. SciNet is funded by: the Canada Foundation for Innovation; the Government of Ontario; Ontario Research Fund - Research Excellence; and the University of Toronto. Details of the MIST computational system are given at: \url{https://docs.scinet.utoronto.ca/index.php/Mist}. All computations for the experiments performed herein were carried out on a single NVIDIA V100 SMX2 32Gb graphics processing unit (GPU). \\

Data preprocessing (i.e. tokenization, and token to BIO-tag alignment, etc.) were performed using base R string processing functions. Transformer models were fine-tuned and evaluated in Python using the Huggingface Transformer library (\url{https://huggingface.co/transformers/}). \\

\section{Results}

\subsection{Description of i2b2 Sample}

The i2b2/UTHealth 2014 PHI de-identification corpus contains N=1,223 clinical notes. The corpus contains N=765,517 tokens. There are N=600 clinical notes in the training sample (N=372,078 tokens), N=137 in the validation sample (N=90,532), and N=486 clinical notes in the test sample (N=302,957 tokens). There exist 23 unique PHI tags observed in the training sample, and 1 tag denoting all other non-PHI tokens. A breakdown of the count/percentage of different PHI tags observed in each of the train, validation, and test samples is given in Table 3 below.\\

\begin{table}[H]
	\caption{Count/percentage of PHI entities in each of the train, validation and test samples of i2b2/UTHealth 2014 corpus.}
	\centering
    \scalebox{0.95}{
	\begin{tabular}{lccc}
		\toprule
	    & Training Sample & Validation Sample & Test Sample \\
        & Count (Percent) & Count (Percent) & Count (Percent) \\
        \midrule
        Non-PHI & 354986 (95.4192) & 86557 (95.6093) & 288898 (95.3594) \\
        DATE & 6116 (1.6440) & 1526 (1.6856) & 5083 (1.6778) \\
        DOCTOR & 3791 (1.0190) & 805 (0.8892) & 3027 (0.9992) \\
        HOSPITAL & 1780 (0.4785) & 404 (0.4463) & 1407 (0.4644) \\
        PATIENT & 1513 (0.4067) & 313 (0.3457) & 1256 (0.4146) \\
        AGE & 911 (0.2449) & 251 (0.2773) & 712 (0.235) \\
        STREET & 495 (0.1331) & 83 (0.0917) & 393 (0.1297) \\
        MEDICALRECORD & 456 (0.1226) & 118 (0.1303) & 412 (0.1360) \\
        CITY & 349 (0.0938) & 72 (0.0795) & 316 (0.1043) \\
        PROFESSION & 320 (0.0860) & 81 (0.0895) & 302 (0.0997) \\
        PHONE & 270 (0.0726) & 69 (0.0762) & 255 (0.0842) \\
        STATE & 239 (0.0642) & 43 (0.0475) & 197 (0.0650) \\
        USERNAME & 204 (0.0548) &  53 (0.0585) & 91 (0.0300) \\
        IDNUM & 180 (0.0484) & 65 (0.0718) & 191 (0.0630) \\
        ORGANIZATION & 166 (0.0446) & 51 (0.0563) & 134 (0.0442) \\
        ZIP & 156 (0.0419) & 26 (0.0287) & 132 (0.0436) \\
        COUNTRY & 64 (0.0172) & 6 (0.0066) & 123 (0.0406) \\
        LOCATION-OTHER & 10 (0.0027) & 2 (0.0022) & 15 (0.0050) \\
        FAX & 9 (0.0024) & 0 (0) & 2 (0.0007) \\
        URL & 6 (0.0016) & 0 (0) & 0 (0) \\
        EMAIL & 3 (0.0008) & 1 (0.0011) & 1 (0.0003) \\
        HEALTHPLAN & 2 (0.0005) & 0 (0) & 0 (0) \\
        BIOID & 1 (0.0003) & 0 (0) & 0 (0) \\
        DEVICE & 1 (0.0003) & 6 (0.0066) & 10 (0.0033) \\
    \bottomrule
	\end{tabular}
	}
	\label{tab:table3}
\end{table}

\subsection{Optimal Hyper-Parameter Configurations}

When training/fine-tuning various transformer architectures we chose the maximal batch size such that the training dataset/batch is capable of fitting into the GPU hardware. We search over a hyper-parameter grid (see methods section) for a best-performing (“optimal”) transformer architecture within a model class. We tune models over the following hyper-parameters: number of training/fine-tuning epochs, learning rate, and weight decay. Below we present the optimal hyper-parameter configurations for each of the model classes being compared in this study (where the hyper-parameter grid for a given model class is given in Table 2 above). \\

\begin{table}[H]
	\caption{Hyper-parameters for transformer model architectures resulting in the best/optimal in-class performance, as measured using the i2b2/UTHealth 2014 de-id validation dataset.}
	\centering
    \scalebox{0.95}{
	\begin{tabular}{lcccc}
		\toprule
	    Model & Batch Size & Number Epochs & Learning Rate & Weight Decay \\
	    \midrule
        Roberta-Large & 12 & 75 & 0.00001 & 0.01 \\
        Albert-XXLarge & 6 & 20 & 0.0001 & 0.0 \\
        Roberta-Base & 35 & 50 & 0.0001 & 0.01 \\
        Bert-Large & 12 & 40 & 0.0001 & 0.025 \\
        Albert-Base & 35 & 40 & 0.0001 & 0.01 \\
        Bert-Base & 35 & 50 & 0.0001 & 0.01 \\
        \bottomrule
	\end{tabular}
	}
	\label{tab:table4}
\end{table}

\subsection{Overall Model Performance}

In this section we investigate/compare models in terms of operating characteristics on training, validation, and test sets. In general, training set performance exceeds that of validation/test set performance (suggesting some over-fitting may be occurring). Validation/test set performance are similar both in terms of absolute model performance (i.e. report recall, precision, F1-score and overall accuracy), as well as rank ordering of optimally performing models. Larger models (within a class/architecture) perform better than smaller models. There exists some variation in model performance (even within best-in-class models): all models achieve strong accuracy (>99\%); however, recall/precision/F1-score vary within a range of 0.93 – 0.97 depending on best model in a given class.

\begin{table}[H]
	\caption{Precision (PPV), Recall (Sensitivity), F1-Score, and Overall Accuracy of best performing transformer models within a model class, evaluated on training/validation/test datasets.}
	\centering
    \scalebox{0.70}{
	\begin{tabular}{lcccccccccccc}
		\toprule
	& Train & Train & Train & Train & Validation & Validation & Validation & Validation & Test & Test & Test & Test \\
	Model & Precision & Recall & F1-Score & Accuracy &	Precision & Recall & F1-Score & Accuracy & Precision &	Recall & F1-Score & Accuracy \\ 
	\midrule
    Roberta-Large &	0.9997 & 0.9996 & 0.9996 & 1.0000 & 0.9747 & 0.9737 & 0.9742 & 0.9972 & 0.9669 & 0.9681 & 0.9675 &	0.9967 \\
    Albert-XXLarge & 0.9999 & 0.9996 & 0.9997 & 1.0000 & 0.9703 & 0.9643 & 0.9673 & 0.9968 & 0.9662 & 0.9627 &	0.9644 & 0.9965 \\
    Roberta-Base & 1.0000 & 0.9995 & 0.9998 & 1.0000 & 0.9661 & 0.9600 & 0.9630 & 0.9961 & 0.9510 & 0.9533 & 0.9522 &	0.9952 \\
    Bert-Large & 1.0000 & 0.9998 & 0.9999 & 1.0000 & 0.9627 &	0.9598 & 0.9612 & 0.9961 & 0.9553 & 0.9534 & 0.9543 &	0.9956 \\
    Albert-Base & 0.9997 & 0.9999 & 0.9998 & 1.0000 & 0.9496 &	0.9470 & 0.9483 & 0.9949 & 0.9387 & 0.9385 & 0.9386 &	0.9939 \\
    Bert-Base & 1.0000 & 0.9992 & 0.9996 & 1.0000 & 0.9470 & 	0.9460 & 0.9465 & 0.9949 & 0.9380 & 0.9440 & 0.9410 &	0.9946 \\
    \bottomrule
	\end{tabular}
	}
	\label{tab:table5}
\end{table}

\subsection{Performance on Various PHI Sub-Categories}

Transformer model classes/architectures (following hyper-parameter grid search to identify optimally performing models) all illustrate strong performance with respect to overall detection of PHI tokens (accuracy>99\%; F1-score>93\%). However, PHI specific operating characteristics are more variable – suggesting certain PHI-types are more difficult than others for the selected transformer model architectures to identify. For example, figure 3 illustrates that transformer architectures perform well on most PHI-types; however struggle to some extent at identifying professions, organizations, ages, and certain locations.

\begin{figure}[H]
	\centering
	\includegraphics[width=0.85\textwidth]{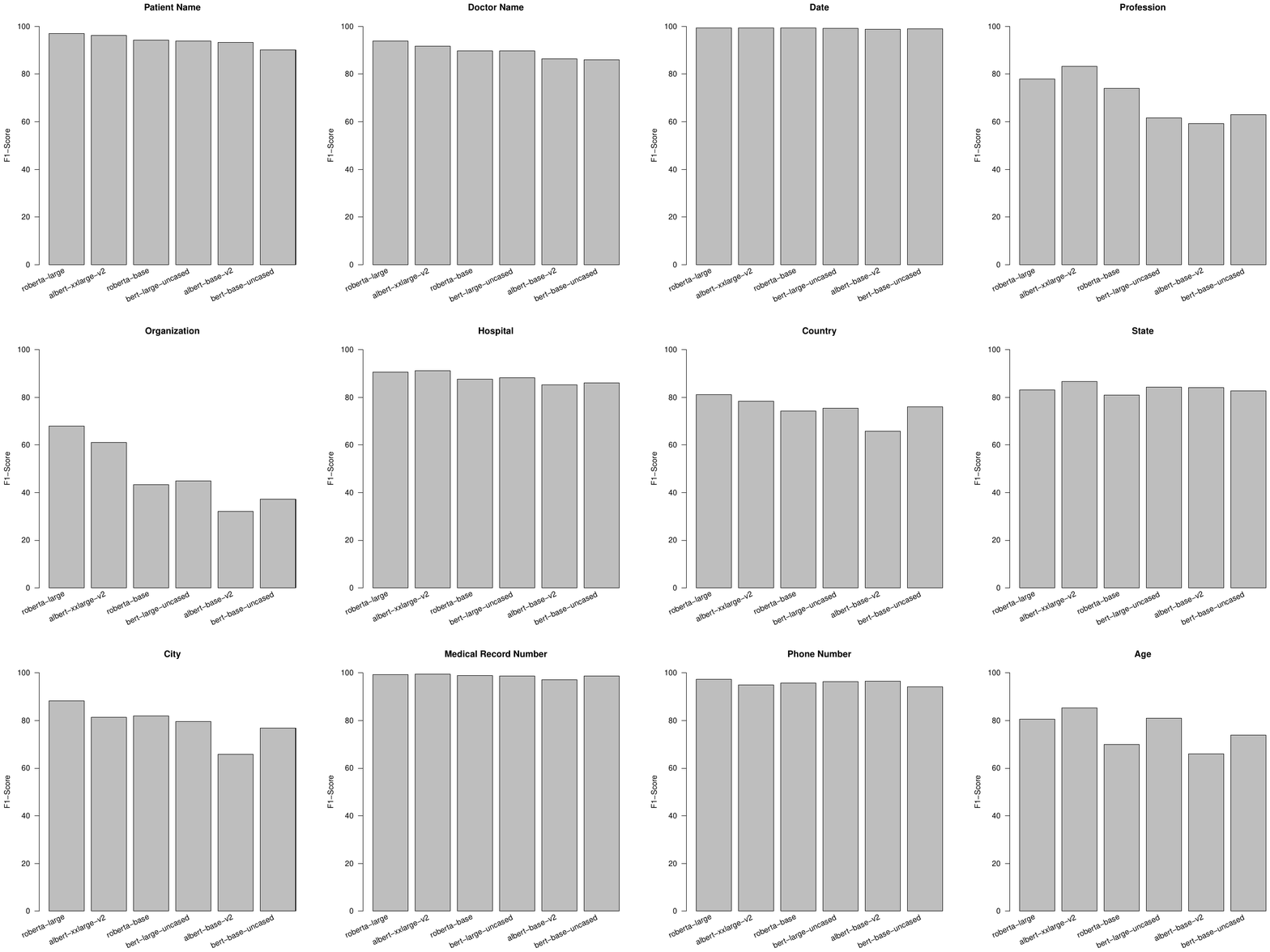}
	\caption{Bar-plot comparing F1-scores across optimally chosen transformer model architectures over a variety of PHI-types.}
	\label{fig:fig3}
\end{figure}

\subsection{Model Size and Model Fine-Tuning Time}

The transformer model architectures being compared in this study vary in terms of size; which subsequently impacts fine-tuning time (and downstream prediction times). Below Table 6 compares chosen transformer model architectures in terms of fine-tuning/throughput times. We observe that larger models are associated with longer training times, and slower overall training throughput. 
 
\begin{table}[H]
	\caption{Training or fine-tuning time.}
	\centering
    \scalebox{0.95}{
	\begin{tabular}{lcccccccccccc}
		\toprule
        Model & Samples Per Second & Steps Per Second \\
        \midrule
        Roberta-Large & 21.073 & 1.846 \\
        Albert-XXLarge & 4.388 & 0.737 \\
        Roberta-Base & 40.864 & 1.193 \\
        Bert-Large & 21.319	& 1.867 \\
        Albert-Base & 38.417 & 1.122 \\
        Bert-Base & 40.918 & 1.195 \\
    \bottomrule
	\end{tabular}
	}
	\label{tab:table6}
\end{table}

\section{Discussion}

As of writing, transformer model architectures represent the state of art in many applied NLP tasks, and increasingly this is true with respect to de-identification of PHI from clinical text data (a type of clinical named entity recognition problem). In this study we comparatively evaluated several transformer model architectures on the i2b2/UTHealth 2014 clinical de-identification corpus and found that after suitable hyper-parameter tuning the ROBERTA-large transformer model architecture performed exceptionally well (96.7\% recall and 96.7\% precision). Other tuned/optimized transformer model architectures also performed well (>99\% accuracy, and 93\%-96\% precision/recall); however, the performance of the ROBERTA-large model is approximately 1\% better than the second-best performing model (in this case ALBERT-xxlarge).\\

After extensive hyper-parameter tuning (over a reasonable sized hyper-parameter grid, including number training epochs, learning rate and weight decay) we observe that all models perform reasonably well at identifying PHI within the i2b2/UTHealth 2014 corpus. Optimally tuned models within a particular transformer model class/architecture all achieve >99\% accuracy, and >93\% recall/precision (see Table 5 above). When performing the model search certain hyper-parameter configurations resulted in inferior performance. Hence, while transformer model architectures offer great promise for clinical text de-identification tasks, off-the-shelf models (with respect to default hyper-parameter configurations) may perform sub-optimally, and hyper-parameter tuning is encouraged (if computationally feasible). \\

Few consistent trends emerge with respect to optimal hyper-configuration options across model classes. With respect to the best performing model in this study (ROBERTA-large), we train/fine-tune for 75 epochs, at a learning rate of 0.00001, with the weight decay (regularization) parameter set to 0.01. Compared to other studies this seems to be a larger number of training epochs than expected. For other models we observe optimal performance at a smaller number of training epochs and using a lower weight decay parameter (0.025, 0.01, 0.0). \\

One consistent trend we observe, is that large models all seem to perform better than their base-sized alternative. That is, ROBERTA-large performance exceeds ROBERTA-base; BERT-large performance exceeds BERT-base; and lastly, ALBERT-xxlarge performance exceeds that of ALBERT-base. Hence, transformer model size does seem to matter; and to the extent that ones GPU memory permits, one should attempt to use large models compared to their base alternatives. Adjusting training/fine-tuning batch sizes is a trick which can be used to help ensure models can be fit on available computational hardware. A trade-off however is that models trained on smaller batch sizes may take longer to fit. Further, when using larger models for downstream prediction problems, feeding out-of-sample examples/data through the model can also be computationally more expensive. \\

\subsection{Research on Clinical Text De-identification}

There has been a steady growth of research conducted on de-identification of clinical text; much of which dates back to publication of the HIPAA guidelines regarding safe/respectful use of sensitive patient health records for research and quality improvement purposes. Research related to clinical text de-identification relies on the availability of annotated training/evaluation corpora, few of which are publicly available to this date. As such, much progress in the field has been focused on a few coordinated efforts to meticulously acquire, annotate, and publicly release small de-identified clinical text datasets which can be used as part of (potentially cross-institutional) data scientific challenges. \\

Oftentimes clinical document collections pertain to a specific clinical domain and are gathered from a single clinical institution; creating a potential issue with generalizability/transportability of learned systems across domains/setting. A selection of publicly available data resources for clinical text de-identification research are described in: \\

\begin{itemize}
 \item \citet{douglass2004computer}: MIMIC-II ICU nursing note corpus.
 \item \citet{uzuner2007evaluating}: i2b2 2006 discharge summary corpus.
 \item \citet{deleger2014preparing}: Cincinnati Children’s Hospital corpus.
 \item \citet{stubbs2015annotating}: i2b2 2014 diabetic inpatient note corpus. 
\end{itemize}

Methodologically, we have seen incremental progress in the field since the advent of HIPAA safe harbour. At least two (systematic) reviews have been published regarding clinical text de-identification research. The review of \citet{meystre2010automatic} discuss methods related to 1) pattern matching, expert derived dictionaries/gazetteers/lexicons, and the application of regular expressions for identification of PHI in clinical notes; and 2) traditional machine learning classifiers (e.g. CRF classifiers; \citep{liu2015automatic}) using manually coded lexical-, morphological-, semantic-, and syntactic-features. Many impressive dictionary or rule-based systems have been implemented; for example, the deid- \citep{douglass2004computer}, Scrubber- \citep{aberdeen2010mitre} and Philter-system \citep{norgeot2020protected}. More recently the review of \citet{yogarajan2020review} discuss advances in the field. Current state of the art models for most clinical text de-identification challenges rely on deep learning architectures; including: recurrent neural networks \citep{dernoncourt2017identification}, long short term memory network, and gated recurrent unit networks \citep{ahmed2020identification} and transformers \citep{johnson2020deidentification, murugadoss2021building}. Hybrid models and ensemble models remain an interesting area for research. \\

\subsection{Limitations and Future Work}

In this study we observed that large transformer models architectures (particularly ROBERTA-large) performed optimally at identifying PHI in the i2b2/UTHealth 2014 corpus. While this finding is true with respect to a particular clinical PHI de-identification corpus; it would be interesting future work to investigate whether ROBERTA-large performs similarly well on other clinical de-identification datasets.\\

Along the lines of training/fine-tuning transformer model architectures on different clinical de-identification corpora, one may want to consider curating a single corpus which combines the clinical text data and tag sets from each of the aforementioned open source clinical de-identification corpora. Each particular corpus has been curated in different clinical settings (i.e. the datasets are gathered from different medical sub-disciplines); hence, this approach to training sample curation may facilitate greater generalizability/transportability of learned models to downstream tasks \citep{yang2019study}. \\
 
The i2b2/UTHealth 2014 clinical corpus is based on diabetic inpatient notes (possibly amongst patient experiencing comorbid cardiovascular sequalae), hence downstream utility of models trained on this corpus may be narrow. For example, it is not clear whether identified models would perform well under possible domain adaptation if applied, say, to primary care clinical notes, emergency department discharge notes, narrative psychiatric assessments, pathology reports, or imaging/radiology consult letters.\\

We have considered a narrow subset of possible transformer models (BERT, ROBERTA, and ALBERT models only), future work could consider fine tuning the Bio-BERT \citep{lee2020biobert} or Sci-BERT \citep{beltagy2019scibert} class of transformer models (n.b. these have been trained over PubMed or other scientific text corpora, rather than basic English text) \citep{johnson2020deidentification}. Larger comparative evaluations which consider a wider class of transformer models may generate insightful interviews: e.g. XLNet \citep{yang2019xlnet} and other models \citep{kalyan2021ammu, kalyan2021ammus}.\\

Lastly, it would be interesting to investigate whether hybrid models (e.g. transformer models coupled with simple REGEX/pattern-matching inspired rules) and/or ensembles of transformers may empirically improve performance with respect to identification of PHI entities. \\

\section{Conclusions}
In this study we comparatively evaluated several transformer architectures (i.e. BERT, ROBERTA, and ALBERT) for clinical text de-identification using the i2b2/UTHealth 2014 corpus. After suitable fine-tuning over extensive hyper-parameter grids, all transformer model classes performed exceptionally well (i.e. >99\% accuracy, >93\% precision/recall). We observe that ROBERTA-large was the best performing model (>99\% accuracy; 96.7\% precision/recall). Overall, we observe that transformer models architectures (after suitable hyper-parameter optimization) offer a satisfactory solution for the clinical text de-identification problem; and could be readily adopted in clinical scenarios where clinicians/researchers are looking to use de-identified clinical text data to facilitate quality improvement and enhanced patient care.  

\newpage

\bibliographystyle{unsrtnat}
\bibliography{references}  

\clearpage

\appendix

 \newpage%
 \renewcommand{\thesection}{\Alph{section}}
 
 \section{First Appendix}
  BIO-tags used in our analysis of the i2b2/UTHealth 2014 de-identification dataset.

\begin{enumerate}
\item B-AGE
\item B-BIOID
\item B-CITY
\item B-COUNTRY
\item B-DATE
\item B-DEVICE
\item B-DOCTOR
\item B-EMAIL
\item B-FAX
\item B-HEALTHPLAN
\item B-HOSPITAL
\item B-IDNUM
\item B-LOCATION-OTHER
\item B-MEDICALRECORD
\item B-ORGANIZATION
\item B-PATIENT
\item B-PHONE
\item B-PROFESSION
\item B-STATE
\item B-URL
\item B-USERNAME
\item B-ZIP
\item I-AGE
\item I-CITY
\item I-COUNTRY
\item I-DATE
\item I-DOCTOR
\item I-FAX
\item I-HEALTHPLAN
\item I-HOSPITAL
\item I-IDNUM
\item I-LOCATION-OTHER
\item I-MEDICALRECORD
\item I-ORGANIZATION
\item I-PATIENT
\item I-PHONE
\item I-PROFESSION
\item I-STATE
\item I-STREET
\item I-URL
\item Outside/Non-PHI
\end{enumerate}
\newpage

  \section{Second Appendix}
  
  \begin{center}
\begin{longtable}{|l|c|}
    \caption{BERT-base model description and parameter counts.}\\
    \hline
    \textbf{Module} & \textbf{Number Parameters}  \\
    \hline
    \endfirsthead
    \multicolumn{2}{c}%
    {\tablename\ \thetable\ -- \textit{Continued from previous page}} \\
    \hline
    \textbf{Module} & \textbf{Number Parameters} \\
    \hline
    \endhead
    \hline \multicolumn{2}{r}{\textit{Continued on next page}} \\
    \endfoot
    \hline
    \endlastfoot

       bert.embeddings.word.embeddings.weight & 23440896 \\
       bert.embeddings.position.embeddings.weight & 393216 \\
       bert.embeddings.token.type.embeddings.weight & 1536 \\ bert.embeddings.LayerNorm.weight & 768 \\
       bert.embeddings.LayerNorm.bias & 768 \\
       bert.encoder.layer.0.attention.self.query.weight &   589824 \\
       bert.encoder.layer.0.attention.self.query.bias & 768 \\
       bert.encoder.layer.0.attention.self.key.weight & 589824 \\
\bottomrule
	\end{longtable}
\end{center}
  
  \newpage

  \section{Third Appendix}
  
  \begin{center}
  \begin{longtable}{|l|c|}
    \caption{BERT-large model description and parameter counts.}\\
    \hline
    \textbf{Module} & \textbf{Number Parameters}  \\
    \hline
    \endfirsthead
    \multicolumn{2}{c}%
    {\tablename\ \thetable\ -- \textit{Continued from previous page}} \\
    \hline
    \textbf{Module} & \textbf{Number Parameters} \\
    \hline
    \endhead
    \hline \multicolumn{2}{r}{\textit{Continued on next page}} \\
    \endfoot
    \hline
    \endlastfoot
       
    bert.embeddings.word.embeddings.weight & 31254528 \\
bert.embeddings.position.embeddings.weight & 524288 \\
bert.embeddings.token.type.embeddings.weight & 2048 \\
bert.embeddings.LayerNorm.weight & 1024 \\
bert.embeddings.LayerNorm.bias & 1024 \\

bert.encoder.layer.0.attention.self.query.weight & 1048576 \\
bert.encoder.layer.0.attention.self.query.bias & 1024 \\
bert.encoder.layer.0.attention.self.key.weight & 1048576 \\
\bottomrule
	\end{longtable}
\end{center}
  
  \newpage

  \section{Fourth Appendix}
  
    \begin{center}
  \begin{longtable}{|l|c|}
    \caption{ROBERTA-base model description and parameter counts.}\\
    \hline
    \textbf{Module} & \textbf{Number Parameters}  \\
    \hline
    \endfirsthead
    \multicolumn{2}{c}%
    {\tablename\ \thetable\ -- \textit{Continued from previous page}} \\
    \hline
    \textbf{Module} & \textbf{Number Parameters} \\
    \hline
    \endhead
    \hline \multicolumn{2}{r}{\textit{Continued on next page}} \\
    \endfoot
    \hline
    \endlastfoot
       
    roberta.embeddings.word.embeddings.weight & 38603520 \\
roberta.embeddings.position.embeddings.weight & 394752 \\
roberta.embeddings.token.type.embeddings.weight & 768 \\
roberta.embeddings.LayerNorm.weight & 768 \\
roberta.embeddings.LayerNorm.bias & 768 \\

roberta.encoder.layer.0.attention.self.query.weight & 589824 \\
roberta.encoder.layer.0.attention.self.query.bias & 768 \\
roberta.encoder.layer.0.attention.self.key.weight & 589824 \\
\bottomrule
	\end{longtable}
\end{center}

  \newpage

  \section{Fifth Appendix}
  
      \begin{center}
  \begin{longtable}{|l|c|}
    \caption{ROBERTA-large model description and parameter counts.}\\
    \hline
    \textbf{Module} & \textbf{Number Parameters}  \\
    \hline
    \endfirsthead
    \multicolumn{2}{c}%
    {\tablename\ \thetable\ -- \textit{Continued from previous page}} \\
    \hline
    \textbf{Module} & \textbf{Number Parameters} \\
    \hline
    \endhead
    \hline \multicolumn{2}{r}{\textit{Continued on next page}} \\
    \endfoot
    \hline
    \endlastfoot

roberta.embeddings.word.embeddings.weight & 51471360 \\
roberta.embeddings.position.embeddings.weight & 526336 \\
roberta.embeddings.token.type.embeddings.weight & 1024 \\
roberta.embeddings.LayerNorm.weight & 1024 \\
roberta.embeddings.LayerNorm.bias & 1024 \\

roberta.encoder.layer.0.attention.self.query.weight & 1048576 \\
roberta.encoder.layer.0.attention.self.query.bias & 1024 \\
roberta.encoder.layer.0.attention.self.key.weight & 1048576 \\
\bottomrule
	\end{longtable}
\end{center}

  \newpage

  \section{Sixth Appendix}
  
        \begin{center}
  \begin{longtable}{|l|c|}
    \caption{ALBERT-base model description and parameter counts.}\\
    \hline
    \textbf{Module} & \textbf{Number Parameters}  \\
    \hline
    \endfirsthead
    \multicolumn{2}{c}%
    {\tablename\ \thetable\ -- \textit{Continued from previous page}} \\
    \hline
    \textbf{Module} & \textbf{Number Parameters} \\
    \hline
    \endhead
    \hline \multicolumn{2}{r}{\textit{Continued on next page}} \\
    \endfoot
    \hline
    \endlastfoot
       
albert.embeddings.word.embeddings.weight & 3840000 \\
albert.embeddings.position.embeddings.weight & 65536 \\
albert.embeddings.token.type.embeddings.weight & 256 \\
albert.embeddings.LayerNorm.weight & 128 \\
albert.embeddings.LayerNorm.bias & 128 \\
albert.encoder.embedding.hidden.mapping.in.weight & 98304 \\
albert.encoder.embedding.hidden.mapping.in.bias & 768 \\
albert.encoder.albert.layer.groups.0.albert.layers.0.full.layer.layer.norm.weight & 768 \\
\bottomrule
	\end{longtable}
\end{center}

  \newpage

  \section{Seventh Appendix}
  
          \begin{center}
  \begin{longtable}{|l|c|}
    \caption{ALBERT-xxlarge model description and parameter counts.}\\
    \hline
    \textbf{Module} & \textbf{Number Parameters}  \\
    \hline
    \endfirsthead
    \multicolumn{2}{c}%
    {\tablename\ \thetable\ -- \textit{Continued from previous page}} \\
    \hline
    \textbf{Module} & \textbf{Number Parameters} \\
    \hline
    \endhead
    \hline \multicolumn{2}{r}{\textit{Continued on next page}} \\
    \endfoot
    \hline
    \endlastfoot

albert.embeddings.word.embeddings.weight & 3840000 \\
albert.embeddings.position.embeddings.weight & 65536 \\
albert.embeddings.token.type.embeddings.weight & 256 \\
albert.embeddings.LayerNorm.weight & 128 \\
albert.embeddings.LayerNorm.bias & 128 \\
albert.encoder.embedding.hidden.mapping.in.weight & 524288 \\
albert.encoder.embedding.hidden.mapping.in.bias & 4096 \\
albert.encoder.albert.layer.groups.0.albert.layers.0.full.layer.layer.norm.weight & 4096 \\
\bottomrule
	\end{longtable}
\end{center}

\end{document}